\documentclass{article}
\usepackage{ijcai17}

\usepackage{times}
\usepackage{epsfig}
\usepackage{graphicx}
\usepackage{amsmath}
\usepackage{amssymb}
\usepackage{subfigure}
\usepackage{algorithm}
\usepackage{algorithmic}
\usepackage{placeins}
\usepackage{array}
\usepackage{amsfonts}       % blackboard math symbols
\usepackage{nicefrac}       % compact symbols for 1/2, etc.
\usepackage{microtype}      % microtypography
\usepackage{booktabs}
\usepackage{multirow}

\title{Supervised Adversarial Networks for Image Saliency Detection}
\author{Hengyue Pan and Hui Jiang\\
York University\\
4700 Keele Street, Toronto, Ontario, CA\\
{\tt\small panhy@cse.yorku.ca}, {\tt\small hj@cse.yorku.ca}}
\begin{document}

\maketitle

\begin{abstract}
  In the past few years, Generative Adversarial Network (GAN) became a prevalent research topic. By defining two convolutional neural networks (G-Network and D-Network) and introducing an adversarial procedure between them during the training process, GAN has ability to generate good quality images that look like natural images from a random vector. Besides image generation, GAN may have potential to deal with wide range of real world problems. In this paper, we follow the basic idea of GAN and propose a novel model for image saliency detection, which is called Supervised Adversarial Networks (SAN). Specifically, SAN also trains two models simultaneously: the G-Network takes natural images as inputs and generates corresponding saliency maps (synthetic saliency maps), and the D-Network is trained to determine whether one sample is a synthetic saliency map or ground-truth saliency map. However, different from GAN, the proposed method uses fully supervised learning to learn both G-Network and D-Network by applying class labels of the training set. Moreover, a novel kind of layer call conv-comparison layer is introduced into the D-Network to further improve the saliency performance by forcing the high-level feature of synthetic saliency maps and ground-truthes as similar as possible. Experimental results on Pascal VOC 2012 database show that the SAN model can generate high quality saliency maps for many complicate natural images.
\end{abstract}

\section{Introduction}
\label{Intro}

'Saliency' is originally a definition in the fields of neuroanatomy and psychology. This term denotes a stimulus (from a variety of sources such as images or sounds) that can grab more attention from the observer. For human beings and other organisms, the ability of saliency detection is important since it can help them to find the important resources or potential threats from the surrounding environment. In visual perception, contrast is one important reason of saliency. For example, an object in an image that has different color or texture with its surrounding backgrounds may tend to be recognized as a salient object \cite{schneider1977controlled}.

\begin{figure*}[t]
\begin{center}
    \includegraphics[width=0.9\linewidth]{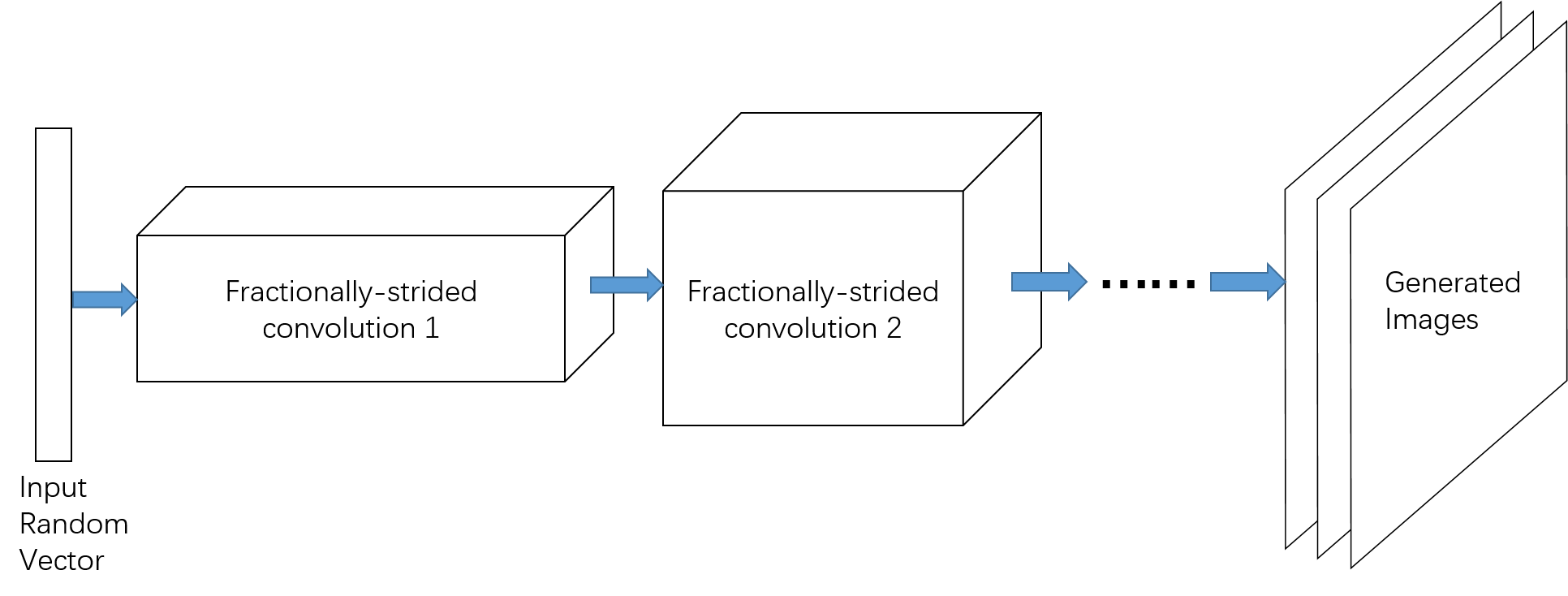}
\end{center}
\caption{The basic structure of the G-Network in GAN. By using fractionally-strided convolutions, the input vector will be converted into several feature maps. The size of feature maps will gradually increase while the number will decrease. Finally the output is the fake images.}
\label{Fig:GAN_GNet}
\end{figure*}

In the domain of computer vision, visual saliency detection is also a well-known research topic, since making computer learn how to find saliency objects (human may also pay more attention on these objects) automatically is important for the development of artificial intelligence. The main object of image saliency detection is to find a saliency map for the input image that can reflect the saliency level of each region \cite{BooleanMap-2013}. The pixel-wised saliency maps can show how likely one given pixel belongs to one of the salient objects \cite{borji2012salient}. Generating such saliency maps automatically has recently raised a great amount of research interest \cite{visual-model-2013} since the saliency maps can help diversity of computer vision tasks like semantic segmentation or object detection \cite{cheng2011global}. Previous researchers have proposed a variety of methods to model the procedure of human attention for image saliency detection \cite{frintrop2010computational}, and those methods can be divided into several classes such as bottom-up methods and deep learning based methods.

\begin{figure*}[t]
\begin{center}
    \includegraphics[width=0.9\linewidth]{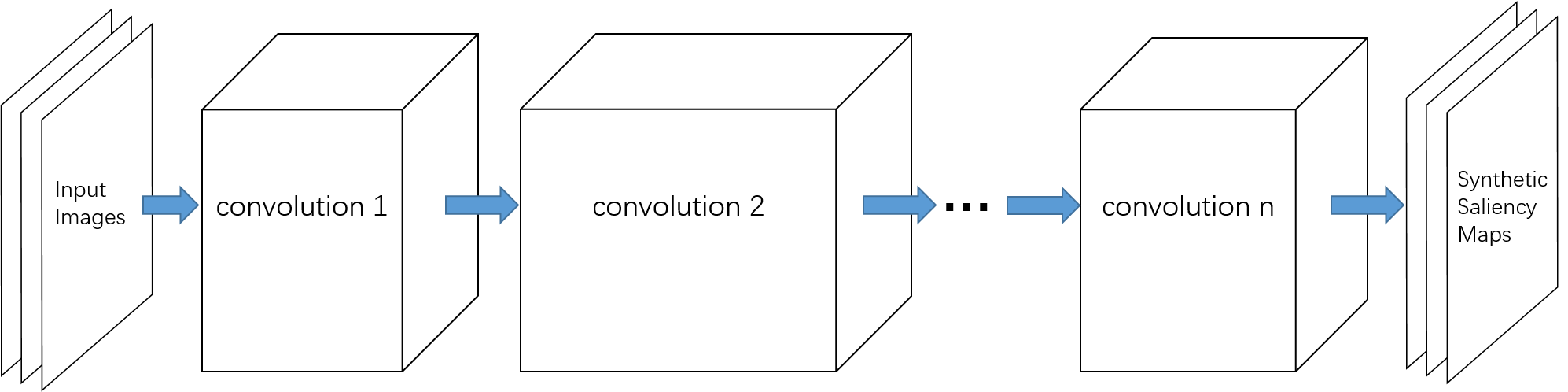}
\end{center}
\caption{The basic structure of the G-Network in SAN. The feature maps' size of all layers are same, while the number of feature maps should firstly increase and then decrease.}
\label{Fig:SAN_GNet}
\end{figure*}

Bottom-up saliency methods are based on the assumption that saliency regions in one image may different with the background. Those methods tend to use low-level features such as color distribution, local/global contrast and texture to generate saliency maps. In \cite{cheng2011global}, a bottom-up saliency method called Region Contrast (RC) is proposed, which considers region contrast and spatial coherence at the same time to achieve good performance. In \cite{fu2013superpixel}, a superpixel level bottom-up saliency method is proposed. This method considers color contrast and color distribution of the superpixels to generate good saliency maps. Moreover, \cite{liu2011learning,riche2012rare,borji2012salient,rahman2015saliency} also provide different kind of bottom-up saliency methods. Even though those bottom-up saliency methods can achieve good performance on some images, it is obvious that simply using low-level features may fail to deal with some complicate images, such as the images with very complex backgrounds or foreground objects. To introduce some global cues or prior information may relieve the problem \cite{harel2006graph,BooleanMap-2013,visual-model-2013}, but those methods still have limitations when dealing with some complex real images.

One possible way to solve the problems of bottom-up methods is to utilize some high-level features to calculate the saliency value. Therefore, some deep learning based saliency methods are proposed, because DNNs and CNNs may have ability to extract more abstract mulit-level features from the images and those different level of features can be used for image saliency detection. \cite{zhao2015saliency} proposed a CNN-based multi-contexts saliency method called multi-context deep saliency. The method considers global context and local context on each superpixel simultaneously, and trained two CNNs to model those features. The multi-level saliency features will finally be combined to generate high quality saliency maps. In \cite{Oxford-cnn-2014}, a CNN back-propagation based saliency framework with a simple objective function is proposed. Even though the performance is not very good, it provides a good idea to do saliency detection, i.e., back-propagating error signals to directly modify the input images and get saliency maps. In \cite{pan2015deep}, Pan et al. extended the saliency framework in \cite{Oxford-cnn-2014} by introducing both original images and masked images as training data and applying image erosion and dilation to improve the raw saliency maps. In \cite{pan2016deep}, a novel objective function is defined to guide the back-propagation process, and the resulting gradients with respect to input images tend to only remove the salient objects and maintain the background regions unchange. Moreover, the method applies superpixel maps and low-level features to further improve the raw saliency maps and finally achieves excellent saliency performance.

Currently, Generative Adversarial Network (GAN)\cite{goodfellow2014generative} becomes a prevalent research topic. Originally, GAN model is designed to do image generation \cite{goodfellow2014generative,che2016mode,zhao2016energy}. To improve the performance of traditional generative model and their unsupervised learning algorithms, GAN introduces two networks, i.e., G-Network and D-Network, and make them 'combat' with each other to improve the performacne. The G-Network tries to generate 'fake' images to cheat the D-Network, while the D-Network is trained to distinguish 'fake' images from real natural images. Experiments show that the adversarial learning process improves the network performance, and many generated images of GAN may look like natural images. Moreover, in \cite{li2016deep}, a GAN based face modification algorithm is proposed, which has ability to modify the input faces, such as change age, gender and expressions. To improve the face quality, this paper proposes to use extra CNNs to introduce identity-aware loss and enhance the visual quality of the generated faces. \cite{arjovsky2017wasserstein} argues that the main problem of GAN is the instability during the training process, and lack of performance measurement. Therefore, it proposes a stable learning method as well as a suitable way to evaluate the quality of generated images for GAN models. In \cite{salimans2016improved} and \cite{odena2016conditional}, semi-supervised GAN model are applied for image generation tasks, which also improves the performance. Recently, some researches present that GAN also have potential to be applied in some other research fields. For instance, in \cite{luc2016semantic}, GAN is used in semantic segmentation, and the class specific label maps as well as a corresponding loss function are applied to adversarial training.

In this paper, we propose a novel model, called 'Supervised Adversarial Network (SAN)', to deal with image saliency detection. This model uses the adversarial feature of GAN, but introduces some new modifications to make it work on saliency tasks. Firstly, we modify the network structure of the G-Network to make it compatible with saliency detection. The G-Network should take natural images as inputs and output the corresponding saliency maps (we call them synthetic saliency maps). Secondly, we apply a novel layer called 'conv-comparison' layer in the D-Network to force the synthetic saliency maps have some identical high-level feature as ground-truth saliency maps. Thirdly, we apply fully-supervised training to provide more precise gradient and relieve the problem of gradient vanishing. After the training, we further do post-processing such as superpixel smoothing and low-level feature refining on synthetic saliency maps to further improve the performance. The experimental results on Pascal VOC 2012 \cite{Everingham10} show that SAN model yields good performance on saliency detection tasks, especially for those relatively complicate images.

\section{Supervised Adversarial Networks}
\label{SAN}

In most of previous methods such as \cite{goodfellow2014generative}, \cite{che2016mode} and \cite{zhao2016energy}, GAN models are designed to generate fake natural images that can 'cheat' the classification CNNs. The unsupervised learning method allows GAN model to use large amount of training data to improve the performance. Comparing with image generation, image saliency detection is a very different task, which has rigid ground-truthes that can be used to measure the performance. In this section, we will discuss the proposed Supervised Adversarial Networks (SAN) for image saliency detection.

\subsection{G-Network}

Comparing with GAN model, SAN has a different structure of G-Network. In GAN, the G-Network receives random vectors as inputs, and applied several fractionally-strided convolution layers to expand the input vectors to several square feature maps, and the final outputs are fake images \cite{radford2015unsupervised} (See Figure~\ref{Fig:GAN_GNet} for more details).

By contrast, based on the definition of image saliency detection, the G-Network in SAN requires to use natural images as inputs, and the outputs should be the corresponding saliency maps. Therefore, the G-Network in SAN should use the regular convolution layers instead of the fractionally-strided convolution layers in GAN, and remove the pooling layers to guarantee that the saliency maps have the same size as the input images (See Figure~\ref{Fig:SAN_GNet} for more details). Specifically, in the G-Networks of SAN, every hidden convolutional layer should be followed by one batch normalization layer and one regular ReLU layer. However, following the idea in \cite{radford2015unsupervised}, we do not apply batch normalization to the output layer, and the activation function of the output layer is sigmoid instead of ReLU.

According to Figure~\ref{Fig:SAN_GNet}, the synthetic saliency maps of SAN can have multiple feature maps, since the experiments reflect that this configuration can increase the stability and performance of the network. When we generate multiple dimension saliency maps in practice, the corresponding ground-truthes for D-Network learning should be expanded to the same number of dimensions. And finally we can take average over all dimensions to get the final saliency maps.

\subsection{D-Network}

Generally speaking, the D-Network in SAN has similar structure as its counterpart in GAN, since both of them are designed for classification. Following the configuration in \cite{radford2015unsupervised}, the D-Network also applies batch normalization to most convolution layers, but it will use leaky ReLU \cite{maas2013rectifier} instead of the regular ReLU. Moreover, all pooling layers in the D-Network should be replaced by convolution layers with stride 2 (without non-linear activation function).

In regular GAN model, there are only 2 classes, i.e., real images or fake images. However, this may not be suitable for image saliency detect tasks. \cite{salimans2016improved} and \cite{odena2016conditional} began to consider class labels to improve the performance of image generation. Comparing with image generation, image saliency detection has clear and definite ground-truthes to measure the performance of algorithms, and as a result, to generate higher quality saliency maps, we may need more precise and class specified gradients to update both D-Network and G-Network. Moreover, some preliminary experiments show that using 2 classes on saliency task will make D-Network to achieve nearly $100\%$ classification accuracy very fast, and gradient vanishing will happen much easier when updating the G-Network. Therefore, instead of two classes, we also introduce $L+1$ classes into the SAN model, where $L$ is the class number of the training database, and the extra 1 class denotes the synthetic saliency maps. This makes SAN model can be trained under the fully supervised learning criteria.

Moreover, comparing with GAN model, SAN introduces a new kind of layer to further improve the saliency performance, i.e., conv-comparison layer. During the forward procedure, assuming that we input one ground-truth saliency map $S_g$ and its corresponding synthetic saliency map $S_s$ into the D-Network, then the output of the conv-comparison layer can be denoted as $C_g$ and $C_s$ respectively. One very obvious consideration is: if we force $C_f$ similar as $C_g$, then $S_s$ may also tend to be similar with $S_g$. Here we do not directly compare $S_s$ and $S_g$ since the higher layer in CNNs can extract more abstract features with higher dimensions from the input data, which can provide much more rigid constraint are thus very suitable for the comparison. In the back-propagation process, the conv-comparison layer not only back-propagates the error signals from the upper layers (denote by $E_u$), but also calculates mean square error (MSE) between the $C_g$ and $C_s$ to generate another error signal:

\begin{equation}
\label{eq-convcompare}
E_c = \frac{1}{2} ||C_s - C_g||^2
\end{equation}

Then the back-propagated gradient with respect to the output of the conv-comparison (i.e. $C_s$) layer can be calculated as:

\begin{equation}
\label{eq-error}
\frac{\partial E}{\partial C_s} = (1 - \alpha) \frac{\partial E_u}{\partial C_s} + \alpha \frac{\partial E_c}{\partial C_s}
\end{equation}

where $\alpha$ is used to balance the importance of the two errors. Notice that we may need to normalize the gradient of $E_c$ to make it have the same scale as the gradient of $E_u$. Then the new error signal can be used to update the weights of the conv-comparison layer. In practice, we may apply more than one conv-comparison layers in the D-Network to provide stronger constraint and then further improve performance.

\begin{figure}[t]
\begin{center}
    \includegraphics[width=0.95\linewidth]{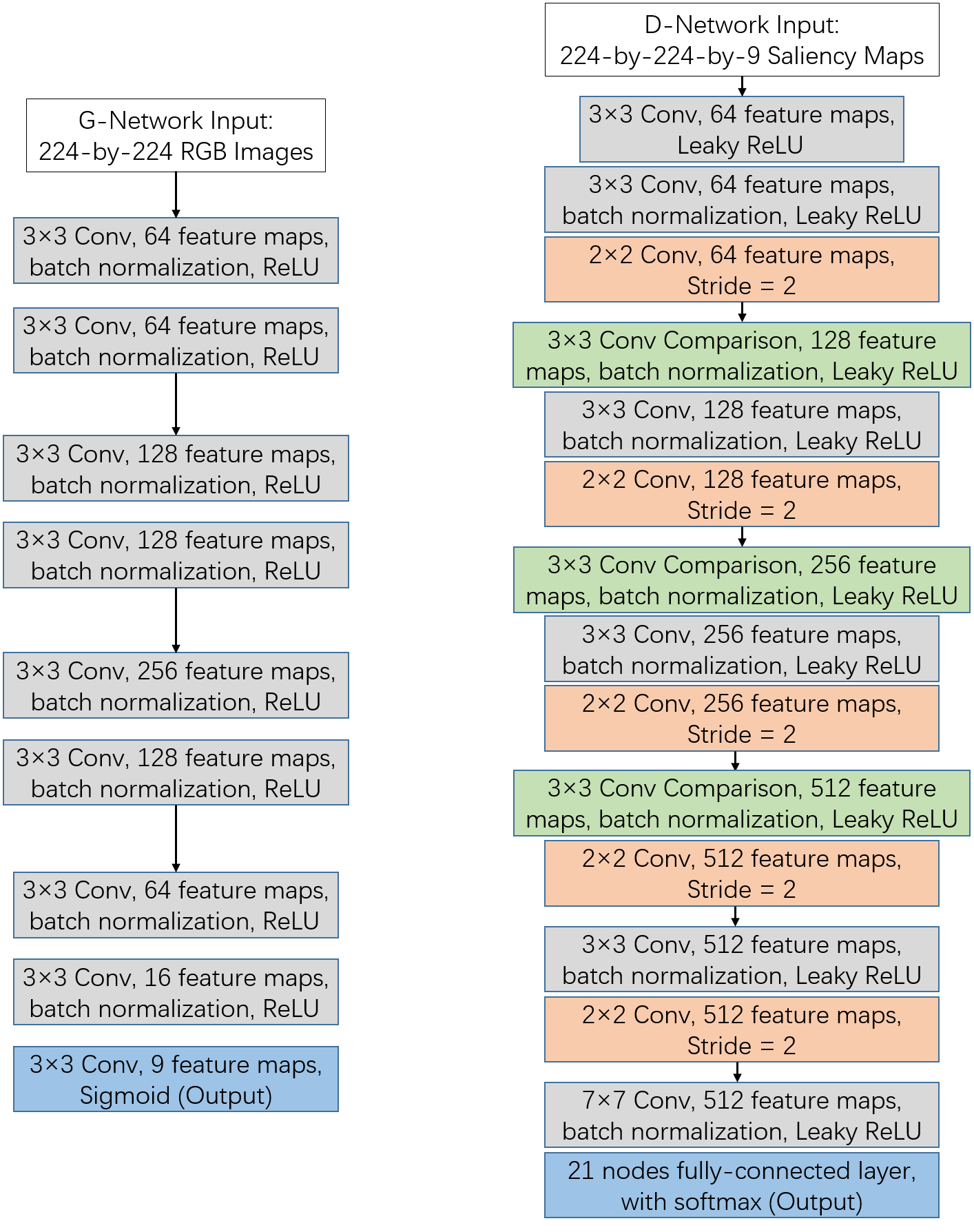}
\end{center}
\caption{The configurations of the G-Network and D-Network in our experiments.}
\label{Fig:GD-Net}
\end{figure}

\subsection{Model Training}

At beginning, all model parameters in SAN should be initialized randomly using the initialization method in \cite{he2015delving}. Then the training process of SAN can be divided into three parts:

\begin{figure*}[htb]
\begin{center}
    \includegraphics[width=1\linewidth]{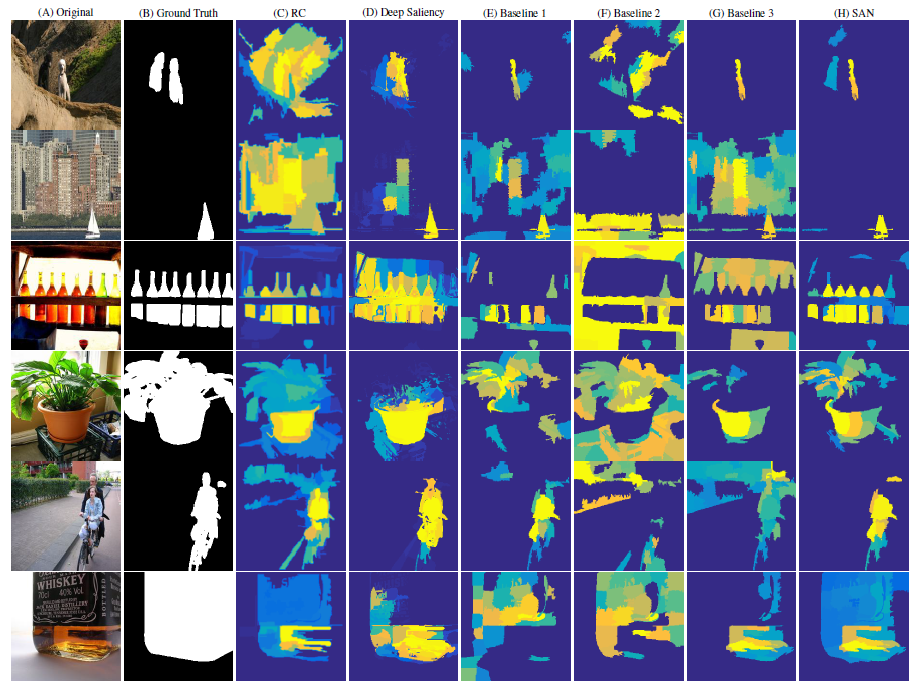}
\end{center}
\caption{Saliency Results of Pascal VOC 2012. (A) original images, (B) ground truth, (C) Region Contrast saliency maps [Cheng {\it et al}., 2011], (D) multi-context deep saliency method [Zhao {\it et al}., 2015], (E) Baseline 1, (F) Baseline 2, (G) Baseline 3. (H) the proposed SAN model. }
\label{Fig:Saliency}
\end{figure*}

% Saliency Results of Pascal VOC 2012. (A) original images, (B) ground truth, (C) Region Contrast saliency maps \cite{cheng2011global}, (D) multi-context deep saliency method \cite{zhao2015saliency}, (E) CNN based method in \cite{pan2016deep}, (F) Baseline 1, (G) Baseline 2, (H) the proposed SAN model.

{\bf (1) saliency maps generation:} In this step, the G-Network works like a pure feed-forward network, which receives training images as inputs and outputs the corresponding synthetic saliency maps. The set of all synthetic saliency maps is denoted by ${\bf S}$.

{\bf (2) updating D-Network:} In this step, we will use ${\bf S}$ as well as the ground-truth saliency maps of the training set (denote by ${\bf G}$) as the training data of the D-Network. The elements in ${\bf G}$ are labelled by using the labels of their corresponding training images (from $1$ to $L$), while all elements in ${\bf S}$ will be labelled as $L+1$, which denotes synthetic saliency maps. By using the pairs of training data and labels, the D-Network can be trained supervisingly via regular error back-propagation algorithm. The well-trained D-Network will achieve a good balance between classification accuracy and providing large enough gradients when updating the G-Network.

{\bf (3) updating G-Network:} After get the well-trained D-Network, we update the G-Network to make it capable to generate better synthetic saliency maps that can 'deceive' the D-Network. This step also can be done by using supervised learning. Specifically, we firstly concatenate the G-Network and D-Network (denote by GD-Network) and use the training images set $I$ as input. During this step all images will be labelled using their original class labels (i.e., from $1$ to $L$). In the forward process, the input signals will firstly pass the G-Network to generate synthetic saliency maps, then the generated saliency maps will pass the D-Network to the output layer. In the backward phase, we fix the weights of D-Network and only update G-Network. During the learning process, the labelling method forces all synthetic saliency maps belong to the corresponding ground-truthes classes. By only updating the G-Network, the distribution of the generated synthetic saliency maps may approach the ground-truthes. In this way, the G-Network may tend to generate higher quality synthetic saliency maps that can make the D-Network recognize them as ground-truth saliency maps.

\subsection{Post-Processing}

After the network training, we can use the G-Network to generate raw saliency maps for validation images. After that, we may use some simple post-processing methods to further improve the quality of raw saliency maps. We firstly filter out some weak signals since it is very possible that those signals are corresponding to background regions. After that, we introduce SLIC superpixels \cite{achanta2012slic} to smooth the raw saliency maps. This operation can weaken some background noises and sharpen the edges of foreground objects. We can further use low-level saliency features mentioned in \cite{fu2013superpixel} to refine the smoothed saliency maps to remove some incorrect saliency regions. Finally the refined saliency maps will be normalized, and we will again filter out weak signals. The details of the post-processing are described in \cite{pan2016deep}.

\section{Experiments}
\label{experi}

We test the presented SAN model on a well-known computer vision database, i.e. Pascal VOC 2012 \cite{Everingham10}, and compare it with several other saliency algorithms. To measure the performance of the selected methods, we use $F_{\beta}$ value as the measurements as \cite{cheng2011global} mentioned. Notice that in our experiments $\beta$ is set to $0.3$ to emphasize the importance of precision.

\subsection{Database}

In image saliency detection and semantic segmentation, Pascal VOC 2012 database \cite{Everingham10} is a classical and also challenging image database. For the saliency and segmentation tasks, this database provides $1464$ training data and $1449$ validation data with their pixel-wised segmentation ground-truthes as well as class labels of all foreground objects. Therefore, Pascal VOC 2012 is suitable for training SAN model and evaluate the algorithm performance.

\subsection{Baseline Methods}

Firstly, we design three baselines to compare with SAN model and demonstrate  the advantages of our configuration. The first one (denote by Baseline 1) is used to reflect the benefits of adversarial learning. Specifically, we remove the D-Network from our model and simply train a G-Network to generate saliency maps. During the training procedure, we firstly initialization the G-Network randomly, and for each training image, we calculate the mean square error (MSE) between the synthetic saliency map and ground-truth to get gradient and update the network. This baseline test will show the saliency performance without adversarial training. The second test (denote by Baseline 2) shares the unsupervised learning criteria with GAN model. Specifically, in this baseline we do not take the class labels of training dataset into account. Instead, similar with GAN, we simply consider 2 classes, i.e., synthetic saliency maps and ground-truth saliency maps, and calculate error signal to update both the D-Network and G-Network based on them. This method will show the importance of the supervised training method introduced in SAN. The Baseline 3 share most of configurations with SAN, and the only different is that we remove all conv-comparison layers from its D-Network. This baseline can reflect the positive influences brought by conv-comparison layers.

Besides the three baselines, we also select two state-of-the-art third-party saliency detection algorithms to compare with the proposed SAN model. The first one is a bottom-up saliency method called Region Contrast (RC) \cite{cheng2011global}. This method considers the global contrast of each superpixel region and introduces spatial constraint to generate saliency maps. The second one is a deep learning based saliency method called multi-context deep saliency, which is proposed by Zhao et al. \cite{zhao2015saliency}. This method introduces two CNNs to model both global and local contexts for each superpixel and generates high quality saliency maps based on those contexts information. % The third one is another CNN based methods presented in \cite{pan2016deep}, which defines a novel objective function and applies back-propagation algorithm to directly modify the input images to generate saliency maps effectively and efficiently.

\subsection{Saliency Results}

In this part we will provide saliency detection results on Pascal VOC 2012. In the following, we use $F_{\beta}$ values and some sample images to evaluate the performance of the proposed SAN model and the other selected baselines. Our computing platform includes Intel Xeon E5-1650 CPU (6 cores), 64 GB memory and Nvidia Geforce TITAN X GPU (12 GB memory). Our algorithms are implemented on MatConvNet platform \cite{MatConvNet-2014}, which is a matlab and CUDA based deep learning toolkit.

\subsubsection{The Selection of Hyperparameters}

In our experiments, the synthetic saliency maps and corresponding ground-truthes have 9 dimensions, and we take average over the 9 dimensions of generated saliency maps to get the final results for evaluation. Since Pascal VOC 2012 has 20 pre-defined classes, thus the output layer of the D-Network needs 21 node to denote all pre-defined classes and one extra class of synthetic saliency maps. In our implementation, the G-Network of SAN has 9 convolution layers, while the D-Network has 15 convolution layers (3 of them are defined as conv-comparison layers) and 1 fully-connected layer (See Figure~\ref{Fig:GD-Net} for more details).

During the learning, we run the training algorithm for 20 iterations. In each iteration, we firstly update the D-Network for 6 epochs, and then update the G-Network for 2 epochs. For the D-Network training, we use 16 mini-batch size. The initial learning rate is 0.0006 and it needs to multiply with 0.98 after every epoch. For the G-Network, we use SGD to do training. The initial learning rate is 0.0001 and the decay rate is also 0.98. We do not use momentum and weight decay in the training process. For the conv-comparison layers in the D-Network, we set $\alpha = 0.8$ to emphasize the gradient of $E_c$.

\subsubsection{Performance}

Table~\ref{table:F_beta} shows the $F_{\beta}$ values of all selected saliency detection algorithms. Comparing with Baseline 1, Baseline 2 and Baseline 3, we can learn that the adversarial learning, fully-supervised training and conv-comparison layers bring about a lot of advantages to saliency results. Moreover, SAN provides better performance than the bottom-up method in \cite{cheng2011global}. Comparing with the CNN based saliency methods in \cite{zhao2015saliency}, the proposed method can also provide slightly better performance.% Comparing with the two CNN based saliency methods in \cite{zhao2015saliency} and \cite{pan2016deep}, the proposed method can also provide comparable performance.

\begin{table}[htb]
\begin{center}
\begin{tabular}{c|c}
\toprule
Methods                         &           $F_\beta$ \\
\midrule
RC \cite{cheng2011global}       &           0.561    \\
Deep Saliency \cite{zhao2015saliency}   &   0.678     \\
% CNN based Method \cite{pan2016deep}     &   0.685     \\
Baseline 1                      &           0.403     \\
Baseline 2                      &           0.382      \\
Baseline 3                      &           0.522      \\
SAN without post-processing     &           0.613     \\
{\bf SAN}                       &         {\bf 0.681}     \\
\bottomrule
\end{tabular}
\end{center}
\caption{The $F_\beta$ value of different saliency methods on Pascal VOC 2012 ($\beta = 0.3$).}
\label{table:F_beta}
\end{table}

Finally, in Figure~\ref{Fig:Saliency}, we provide some examples of the saliency detection results from the Pascal VOC 2012 validation set. From these examples we can see that SAN has ability to solve the image saliency detection tasks in variety of complicate images.

\section{Conclusion}
\label{concl}

In this paper, we have proposed a novel Supervised Adversarial Network (SAN) for image saliency detection. The method relies on the well-known Generative Adversarial Network (GAN) \cite{goodfellow2014generative}, and introduces many modifications to make the model suitable for saliency detection tasks. Specifically, we define a fully-convolution G-Network, which takes images as inputs and outputs corresponding synthetic saliency maps. After that, the synthetic saliency maps and ground-truth saliency maps are used to train the D-Network. We divide the synthetic saliency maps and ground-truthes into $L+1$ classes based on the class labels of the training set, which allows us to train both G-Network and D-Network supervisingly. Moreover, we introduce a novel kind of layer called conv-comparison layer into the D-Network to introduce more constraint to improve the quality of saliency maps. The synthetic saliency maps will be smoothed and refined using SLIC superpixels \cite{achanta2012slic} and low level saliency features \cite{fu2013superpixel}. We have evaluated the performance of the proposed method on Pascal VOC 2012 \cite{Everingham10} database. Experimental results have shown that the introducing of adversarial learning, fully-supervised training and conv-comparison layers can significantly improve the saliency results. Moreover, comparing with the state-of-the-art saliency detection algorithms in \cite{cheng2011global} and \cite{zhao2015saliency}, our proposed method can generate better saliency maps, and also has ability to deal with difficult images.

\bibliographystyle{named}
\bibliography{GAN}

\end{document}